\documentclass[10pt,twocolumn,letterpaper]{article}

\usepackage{wacv}              
\usepackage{pifont}
\newcommand{\cmark}{\ding{51}}
\newcommand{\xmark}{\ding{55}}
\definecolor{wacvblue}{rgb}{0.21,0.49,0.74}
\usepackage[pagebackref,breaklinks,colorlinks,allcolors=wacvblue]{hyperref}
\usepackage{pifont}

\def\wacvPaperID{1344} 
\def\confName{WACV}
\def\confYear{2026}

\title{Eff-GRot: Efficient and Generalizable Rotation Estimation with Transformers}

\author{
Fanis Mathioulakis \quad Gorjan Radevski \quad Tinne Tuytelaars\\
KU Leuven\\
{\tt\small fanis.mathioulakis@kuleuven.be \quad tinne.tuytelaars@kuleuven.be}}

\begin{document}
\maketitle
\begin{abstract}
We introduce Eff-GRot, an approach for efficient and generalizable rotation estimation from RGB images. Given a query image and a set of reference images with known orientations, our method directly predicts the object’s rotation in a single forward pass, without requiring object- or category-specific training. At the core of our framework is a transformer that performs a comparison in the latent space, jointly processing rotation-aware representations from multiple references alongside a query. This design enables a favorable balance between accuracy and computational efficiency while remaining simple, scalable, and fully end-to-end. Experimental results show that Eff-GRot offers a promising direction toward more efficient rotation estimation, particularly in latency-sensitive applications. Code is available at: \url{https://github.com/fmathiou/eff-grot}.

\end{abstract}    
\section{Introduction}
\label{sec:intro}

Estimating the orientation of an object is a fundamental task in computer vision, with applications in robotics, augmented reality, and autonomous driving. While state-of-the-art methods have achieved impressive accuracy and robustness, they often do so at the cost of computational efficiency. In many real-world scenarios—particularly those involving fast-moving objects—reasonably accurate yet fast predictions are preferable to highly precise but slow alternatives. In such cases, slow inference can be detrimental, underscoring the need for efficient and scalable prediction.

In practice, object localization (i.e., translation) can often be reasonably approximated using 2D object crops, which provide x/y position in the image, while depth (z) can be inferred either heuristically (e.g., from known object size) or measured directly using depth sensors. In contrast, rotation estimation remains significantly more challenging. This motivates our decision to target rotation as the core prediction task. By focusing on rotation, we address the more complex aspect of pose estimation, while laying the groundwork for future extensions to full 6D pose in a similarly efficient manner.

\begin{figure}[t]
    \centering
    \includegraphics[width=\columnwidth]{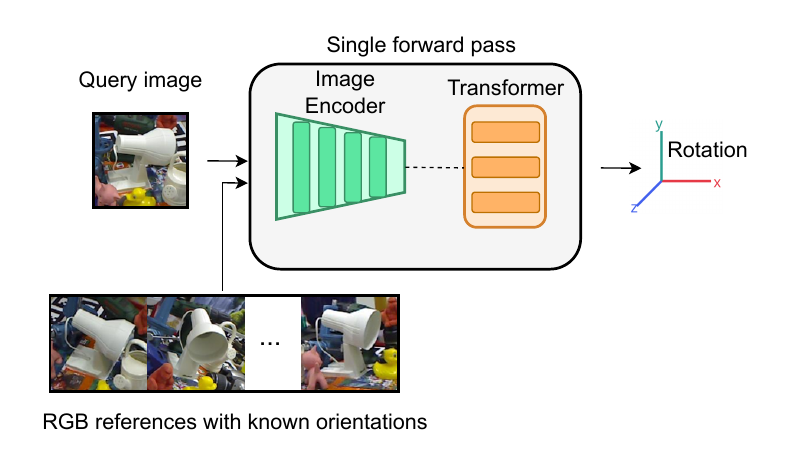}
    \caption{Efficient and generalizable rotation estimation. \mbox{Eff-GRot} estimates the rotation of an unseen object by comparing it to a set of RGB reference images with known orientations. The prediction is made in a single forward pass, enabling fast real-time inference.}
    \label{fig:generalizable_estimation}
\end{figure}

Recent advances in generalizable pose estimation have aimed to move beyond object- or category-specific training~\cite{cai2022ove6d,labbe2022megapose,nguyen2022templates,nguyen2024gigapose,ornek2024foundpose, sun2022onepose, he2022onepose++,liu2022gen6d, pan2024learning, cai2024gs, lee2024mfos}, enabling pose prediction for arbitrary objects using only RGB images and a few posed references. These approaches typically assume access to a set of reference images with known pose information (as shown in Figure~\ref{fig:generalizable_estimation}), allowing them to generalize to novel objects without requiring CAD models. While this model-free paradigm greatly improves flexibility, most existing methods rely on multi-stage pipelines involving correspondence matching~\cite{sun2022onepose, he2022onepose++, lee2024mfos} followed by PnP solvers~\cite{lepetit2009ep}, or iterative optimization~\cite{liu2022gen6d, pan2024learning, cai2024gs}. These components often increase latency and computational cost.

To address the limitations of existing methods, we propose a fully end-to-end and computationally efficient pipeline for generalizable rotation estimation. Our approach predicts the rotation of a novel object in a single forward pass using a transformer that compares a query image to a set of reference images with known rotations. Rather than relying on explicit geometry or appearance matching, the transformer operates in a learned latent space, reasoning about the query’s orientation relative to the references. It leverages global image representations extracted by a frozen encoder, which processes both the query and the reference views. Reference features are enriched with rotation embeddings, while a learnable mask token indicates that the query's rotation should be predicted. This design enables the model to perform 3D reasoning directly from 2D images, eliminating the need for 3D geometry or refinement steps. As a result, our method achieves fast real-time inference speed and is well-suited for latency-sensitive applications.

In summary, our contributions are:
\begin{itemize}
\item We propose an efficient method for generalizable rotation estimation from RGB images, capable of predicting the rotation of novel objects in a single forward pass.
\item We demonstrate that our model offers a strong balance between computational efficiency and accuracy, enabling real-time inference suitable for latency-sensitive applications.
\end{itemize}


\section{Related Work}
\label{sec:formatting}


\paragraph{Instance- and Category-Specific Methods.} Traditional pose estimation methods often assume identical training and test objects—i.e., an instance-specific setting \cite{peng2019pvnet, su2022zebrapose, xiang2017posecnn, wang2021gdr, Wang_2019_CVPR, li2019cdpn, su2015render}. These methods learn canonical poses per object to resolve ambiguities from single-view input. Direct approaches~\cite{li2019cdpn, xiang2017posecnn, su2015render, Wang_2019_CVPR} map images to pose space, while indirect ones~\cite{peng2019pvnet, su2022zebrapose, wang2021gdr} predict 2D–3D correspondences followed by PnP. However, both are limited to seen objects and generalize poorly to novel instances. To address this, category-level approaches \cite{chen2020category, lin2022category, wang2019normalized, chen2020learning} generalize across object classes using shared priors, but still depend on predefined categories and fail with unseen types.

\paragraph{CAD-Based Methods.}
To avoid the limitations of object-specific methods, several works~\cite{cai2022ove6d,labbe2022megapose,nguyen2022templates,nguyen2024gigapose,ornek2024foundpose, zhao2022fusing, lin2024sam} have proposed generalizable pose estimators based on the assumption that 3D CAD models are available for every object at test time. These methods typically render images for a large number of viewpoints for each object and employ a template-based strategy to estimate the pose of a novel object. While these methods can generalize to novel objects, the requirement of having available CAD models limits their applicability for many real-world scenarios.

\paragraph{Reference-Based Methods.} Closest to our work, reference-based methods estimate the pose of novel objects using multi-view reference images with known poses, enabling generalization without requiring 3D models. These approaches can broadly be categorized into two types: correspondence-based and template-based. OnePose and its variants~\cite{sun2022onepose, he2022onepose++} reconstruct a 3D point cloud from posed reference RGB images and match it to 2D keypoints in the query image to obtain 2D–3D correspondences, which are then used to solve for pose via PnP. MFOS~\cite{lee2024mfos} avoids explicit reconstruction by using a transformer to perform patch-level matching and directly regress 2D–3D correspondences.

Template-based methods~\cite{liu2022gen6d, pan2024learning, cai2024gs} avoid explicit correspondence matching altogether. Gen6D~\cite{liu2022gen6d} retrieves a coarse pose through template matching and in-plane rotation regression, followed by refinement using a feature volume. Cas6D~\cite{pan2024learning} builds on this approach with a coarse-to-fine refinement strategy. GSPose~\cite{cai2024gs} reconstructs a 3D object using Gaussian Splatting~\cite{kerbl20233d} and employs differentiable rendering for pose refinement. In contrast, our method is fully end-to-end and directly regresses the object’s rotation in a single forward pass, achieving a favorable balance between efficiency and accuracy.

Finally, the RelPose family of methods~\cite{zhang2022relpose, lin2024relpose++} is designed to jointly recover camera poses from a set of unlabeled images. These methods can be adapted for pose estimation by providing ground-truth poses for a subset of images, allowing RelPose to operate in a reference-based setting—recovering the pose of a target image using labeled reference views. However, they rely on coordinate ascent optimization at inference time, which can be computationally intensive.


\section{Method}
\begin{figure*}[t]
    \centering
    \includegraphics[width=0.8\textwidth]{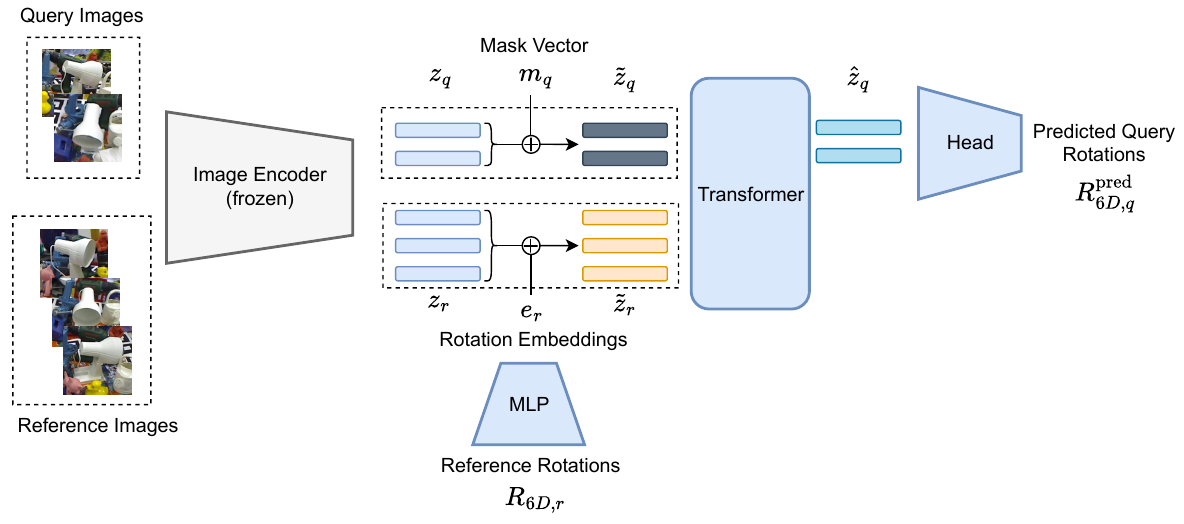}
    \caption{Model overview.
    Eff-GRot takes as input a set of reference images with known rotations and a query image whose rotation needs to be predicted. The encoder processes images from different viewpoints, mapping them to corresponding representations. Learned rotation embeddings are added to reference image representations, while trainable mask vectors are added to query representations. This complete set of representations is fed into a transformer model that outputs updated representations, which are then mapped to the final rotation prediction through a lightweight MLP head.
    }
    \label{fig:architecture}
\end{figure*}

\subsection{Architecture Overview}
An overview of our model can be seen in Figure~\ref{fig:architecture}. \mbox{Eff-GRot} takes as input a set of reference images \(\{I_r^i\}_{i=1}^{N}\) with known rotation information \(\{R_r^i\}_{i=1}^{N}\), and a query image \(I_q\) for which the rotation \(R_q\) must be predicted. We assume that all images are provided as object-centered crops. Each image is encoded into a latent representation using a pretrained encoder \(E\): 

\begin{equation}
    z_r^{i} = E(I_r^{i}),
\end{equation}

where \(z_r^i \in \mathbb{R}^d\) is the mapped feature representation. This encoding enables the model to leverage a compressed yet informative feature space for rotation estimation. 

We experiment with two different encoders: a pretrained Variational Autoencoder (VAE)~\cite{kingma2013auto} from Stable Diffusion v1.5~\cite{rombach2022high}, hereafter referred to as SD-VAE, and a distilled version~\cite{taesd} of it (using the taesd weights as released in the official repository), which we refer to as Distilled VAE.

For representing the rotation, we adopt a 6-dimensional representation~\cite{zhou2019continuity}, composed of the first two columns of the rotation matrix, which has been shown to be well-suited for learning due to its continuity properties. We update the encoded representations of the references by incorporating rotation information into their feature representations. Specifically, we use a small multi-layer perceptron (MLP) to map the 6D rotation encoding \(R^i_{\text{6D}, r} \in \mathbb{R}^6\) to a rotation embedding:

\begin{equation}
    e_r^i = \text{MLP}_{\text{rot}}(R^i_{\text{6D}, r}),
\end{equation}

where \( e_r^i \in \mathbb{R}^d \) is the learned rotation embedding. The final representation for each reference image is then obtained by adding the rotation embedding to its encoded feature representation:

\begin{equation}
    \tilde{z}_r^i = z_r^i + e_r^i.
\end{equation}

This enriched representation captures both the visual appearance and orientation information of the reference images. For the query, we add a continuous learnable mask vector to its representation to indicate that the rotation needs to be predicted:

\begin{equation}
    \tilde{z}_q = z_q + m_q,
\end{equation}

where \( m_q \in \mathbb{R}^d \) is a learnable mask vector. This mask allows the model to differentiate between reference images with known rotations and the query image for which the rotation needs to be estimated.

The entire set, consisting of reference representations with added rotation embeddings and the masked query representation, is provided as input to a transformer model for predicting the query's rotation. By attending to all image representations, the transformer can capture global dependencies and learn rotation relationships.  The transformer processes the input set and produces updated feature representations. Specifically, the updated query embedding at the output of the transformer, denoted as \(\hat{z}_q\), is mapped through a two-layer MLP to a 6-dimensional vector from which the rotation is obtained:

\begin{equation}
    R^{pred}_{\text{6D}, q} = \text{MLP}_{\text{head}}(\hat{z}_q),
\end{equation}

where \(  R_{\text{6D}, q} \in \mathbb{R}^6 \) represents the predicted 6D rotation encoding. The final rotation matrix \( R^{pred}_q \) is then recovered using the Gram-Schmidt orthogonalization process:

\begin{equation}
    R^{pred}_q = \text{Proj}_{\text{SO(3)}}( R^{pred}_{\text{6D}, q}),
\end{equation}

where \( \text{Proj}_{\text{SO(3)}}(\cdot) \) denotes the function that maps the 6D representation to a valid rotation matrix in \( \text{SO(3)} \).

\subsection{Training and Inference}
\label{subsec:training_inference}
For training the model, we keep the VAE encoder frozen and only train the transformer and the two MLPs used for obtaining the rotation embeddings $ e_r^i$ and the output rotation encodings $R_{\text{6D}, q}$ respectively. We apply an L2 loss to the predicted encodings corresponding to the queries:
\begin{equation}
\mathcal{L}_{L2} = \|  R_{\text{6D}, q}^{pred} -  R_{\text{6D}, q}^{gt} \|_2^2
\end{equation}
where $  R_{\text{6D}, q}^{pred}, R_{\text{6D}, q}^{gt} $ are the predicted and ground truth rotation encodings respectively. The entire pipeline is fully end-to-end trainable.

During inference, Eff-GRot requires only the encoded references $\tilde{z}_r^i$ and can predict the rotation of an arbitrary number of query images in a single forward pass. The onboarding stage consists solely of encoding the reference images and does not rely on computationally expensive 3D reconstruction algorithms, such as those used in OnePose~\cite{sun2022onepose}, OnePose++~\cite{he2022onepose++}, or GSPose~\cite{cai2024gs}. This design keeps the entire pipeline simple and efficient.

\paragraph{Rotation Augmentation}  
To improve the generalization of our model, during training, we multiply the rotation matrices of all reference views by a common, randomly sampled rotation matrix \( R \), applied independently for each object in the batch. This prevents the network from associating visual appearance with specific orientations and encourages it to infer the query's rotation by comparing it to the references.

Specifically, let \( R_{oi} \) denote the rotation matrix of reference \( i \) for object \( o \) in the batch. We update the rotation as follows:
\begin{equation}
R'_{oi} = R_{oi}R, \quad \forall i \in S
\end{equation}
where \( S \) is the set of reference indices for object \( o \).

This transformation corresponds to rotating the local reference frame of the object and ensures that the relative rotation among all viewpoints remains consistent.

\section{Experiments}
\label{sec:experiments}

\subsection{Implementation Details}
\label{subsec:implementation}

We use the ViT-Small architecture~\cite{dosovitskiy2020image} for our transformer model, initialized with DINOv2 weights~\cite{oquab2023dinov2}. The model is trained end-to-end using the AdamW optimizer~\cite{loshchilov2017decoupled}, with a learning rate of $10^{-5}$, and batches consisting of 10 objects, each with 94 images (64 references and 30 queries). We use a crop size of $64 \times 64$ for all images. Training takes roughly 30 hours (for 2000 steps) on an NVIDIA RTX 3080 Ti. During training, reference views are randomly sampled for each batch, while during evaluation, we apply farthest point sampling (FPS) to achieve more uniform coverage, following standard practice~\cite{liu2022gen6d, cai2024gs}. To improve generalization to real-world images—where objects often appear in diverse scenes—we augment our training data with backgrounds from the SUN397 dataset~\cite{xiao2010sun}. For evaluation in LINEMOD, each reference image is additionally rotated using five predefined angles within the range of -40 ° to 40 °, increasing coverage and helping the model to better interpolate in the latent space.

\subsection{Experimental Setup}
\label{subsec:experimental_setup}

\paragraph{Datasets.}
For training, we use objects from ShapeNet~\cite{chang2015shapenet} and Google Scanned Objects~\cite{downs2022google}, which provide a large collection of diverse CAD models. Specifically, we use rendered images of approximately 2,000 ShapeNet objects from~\cite{liu2022gen6d}, and 1,023 Google Scanned Objects rendered by~\cite{wang2021ibrnet}.
For evaluation, we use the widely adopted LINEMOD dataset~\cite{hinterstoisser2012model}, which contains real images of 13 objects captured from multiple viewpoints. Following standard practice~\cite{liu2022gen6d, sun2022onepose, he2022onepose++, cai2024gs, pan2024learning}, we sample reference images from the training split and evaluate on the test split. Reference images are cropped using ground-truth bounding boxes, while test image crops are obtained using the off-the-shelf YOLOv5~\cite{jocher2020ultralytics} object detector. We additionally evaluate on 10 categories from the Common Objects in 3D (CO3D) dataset~\cite{reizenstein2021common}, as well as on 388 objects from 10 ShapeNet classes that were excluded from training.

\begin{table}[t]
\centering
\begin{tabular}{@{}lccc@{}} 
\toprule
Method & \multicolumn{3}{c}{Number of references} \\
\cmidrule(lr){2-4}
& 16 & 32 & 64 \\
\midrule
Nearest Ref. & 39.4 & 62.3 & 84.6 \\
Eff-GRot & \textbf{72.7} & \textbf{90.7} & \textbf{94.8} \\
\bottomrule
\end{tabular}
\caption{Results on ShapeNet for different numbers of reference views (ACC @ 15°). Our method significantly outperforms the nearest reference oracle across all reference view settings.}
\label{tab:shapenet_results}
\end{table}

\paragraph{Metrics} To evaluate rotation estimation performance, we use the Acc @ 15\textdegree\ metric, which corresponds to the accuracy of predictions for a 15\textdegree\ threshold where the angular distance is given by the equation:
\begin{equation}
\theta = \cos^{-1} \left( \frac{\text{trace}(R_1^T R_2) - 1}{2} \right)
\end{equation}

\paragraph{Baselines.} We evaluate Eff-GRot across multiple datasets and against a diverse set of baselines to highlight its generalization ability. On ShapeNet, where objects are unseen during training, we compare against a nearest-reference oracle baseline, which selects the reference image closest to the query and copies its rotation. This provides an upper bound for template-based approaches without learning. On LINEMOD, we include comparisons with other generalizable methods that have released both code and pretrained models. Specifically, we compare against OnePose++~\cite{he2022onepose++}, a 2D-3D matching method, and two template-based models: Gen6D~\cite{liu2022gen6d} and the recently proposed GSPose~\cite{cai2024gs}. On CO3D, we include a comparison with RelPose~\cite{zhang2022relpose}.

We omit MFOS from evaluation due to the lack of public code and pretrained models. Although RelPose++ has released code, it does not support reference-based pose estimation, and adapting it to our setting would require significant reimplementation. We instead compare with the original RelPose~\cite{zhang2022relpose}, which provides results and code for reference-based prediction on CO3D.

\begin{table*}[t]
\centering
\renewcommand{\arraystretch}{1.1}
\resizebox{\textwidth}{!}{%
\begin{tabular}{@{}l*{13}{c}cc@{}}
\toprule
Model & ape & benchvise & cam & can & cat & driller & duck & eggbox* & glue* & holepuncher & iron & lamp & phone & Avg. & Subs. Avg. \\
\midrule
\multicolumn{16}{c}{\textit{Low Runtime} ($<$0.1s)} \\
\cmidrule(lr){1-16}
Gen6D* w/o refinement & -- & 70.9 & 81.5 & -- & 72.7 & 64.6 & 56.4 & 78.8 & 74.5 & -- & -- & 79.8 & -- & -- & 72.4 \\
GSPose w/o refinement & 58.2 & 66.2 & 68.5 & 61.1 & 57.0 & 57.8 & 52.1 & 67.0 & 59.9 & 68.9 & 61.8 & 58.8 & 57.6 & 61.2 & 60.9 \\
Eff-GRot & 80.9 & 82.6 & 91.9 & 89.4 & 86.5 & 69.9 & 87.4 & 91.4 & 80.3 & 78.6 & 82.1 & 94.5 & 87.4 & \textbf{84.8} & \textbf{84.5} \\
\midrule
\multicolumn{16}{c}{\textit{High Runtime} ($>$0.1s)} \\
\cmidrule(lr){1-16}
Gen6D* & -- & 88.9 & 95.8 & -- & 95.3 & 78.6 & 79.0 & 98.6 & 95.9 & -- & -- & 94.9 & -- & -- & 90.9 \\
OnePose++ & 57.9 & 96.5 & 99.2 & 98.3 & 81.5 & 92.6 & 86.0 & 85.6 & 58.9 & 88.1 & 94.3 & 99.6 & 85.7 & 86.5 & 87.5 \\
\midrule
\multicolumn{16}{c}{\textit{Very High Runtime} ($\sim$1s)} \\
\cmidrule(lr){1-16}
GSPose & 97.4 & 99.4 & 99.7 & 94.8 & 99.4 & 96.0 & 98.5 & 91.7 & 96.8 & 96.9 & 99.4 & 97.2 & 90.1 & 96.7 & 97.3 \\
\bottomrule
\end{tabular}%
}
\caption{Experimental results on LINEMOD. We evaluate performance on novel objects under the Acc @ 15\textdegree metric for different generalizable methods. * indicates that the model was trained on a subset of LINEMOD, thus the training objects are excluded. For a fair comparison with Gen6D, we provide two different mean values corresponding to all and to a subset of the objects respectively.}
\label{tab:linemod_results}
\end{table*}

\subsection{Comparisons with Baselines}
\label{subsec:comparisons}

\paragraph{Results on ShapeNet.} Table~\ref{tab:shapenet_results} shows quantitative results on novel object categories on the synthetic ShapeNet dataset. 

As expected, the accuracy increases with the number of references, reaching above $90\%$ when using 32 reference views. 
These scores are significantly higher than the oracle baseline based on the nearest reference view. This demonstrates the ability of Eff-GRot to generalize beyond the references, performing meaningful rotation-aware interpolation in the latent space.

\paragraph{Results on LINEMOD.} In Table~\ref{tab:linemod_results} we present quantitative comparisons on rotation estimation performance with other generalizable methods under the Acc @ 15\textdegree metric on LINEMOD using 64 reference views. We split the table in different categories, based on the runtime needed for inference. When excluding the refinement step of Gen6D and GSPose, these methods are in a runtime range similar to that of Eff-GRot (details on runtime are discussed in Section~\ref{subsec:runtime}), yet Eff-GRot systematically obtains higher rotation estimation performance across all objects compared to these methods. In fact, Eff-GRot obtains performance competitive with that of significantly slower methods such as OnePose++ and Gen6D. Compared to OnePose++, our method shows more consistent results, with a clear benefit for objects such as ape and glue, for which OnePose++ seems to struggle. The performance of Eff-GRot is only slightly below that of Gen6D, which was trained on a subset of LINEMOD, so exposed to a smaller domain shift compared to our model trained on synthetic data only. 
Although GSPose achieves the highest overall performance,  this comes at a high computational cost due to the necessity of running an optimization algorithm based on differentiable rendering.

\begin{figure}[tb]
    \centering
    \includegraphics[width=0.7\columnwidth]{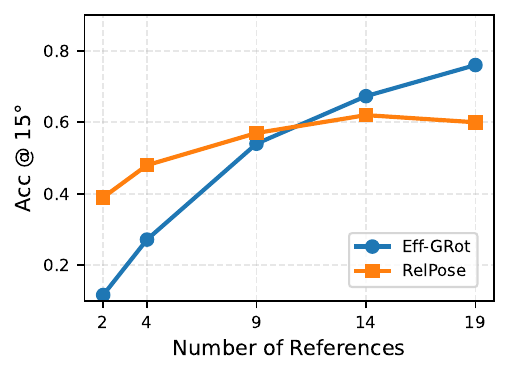}
    \caption{Comparison with RelPose on the CO3D dataset. While RelPose shows diminishing returns beyond 9 reference views, Eff-GRot continues to improve steadily with additional views.}

    \label{fig:relpose}
\end{figure}

\paragraph{Results on CO3D.}
We also compare our method to RelPose on the CO3D dataset. RelPose is designed to jointly recover the camera viewpoints of a set of input images. It can be adapted for rotation estimation by providing ground-truth rotations for a set of reference images and predicting the rotation of a new query image. In Figure~\ref{fig:relpose}, we report accuracy at a 15\textdegree threshold as a function of the number of reference views. RelPose results were approximated from Figure 9 in~\cite{zhang2022relpose}. Eff-GRot achieves performance comparable to RelPose at 9 reference images and surpasses it with more views. Notably, RelPose’s performance shows diminishing returns beyond 9 reference views, while Eff-GRot continues to improve steadily. With 19 reference views, Eff-GRot outperforms RelPose by nearly 20 percentage points (76\% vs. 60\%). Moreover, Eff-GRot is also significantly more efficient: RelPose relies on coordinate ascent optimization for rotation inference, resulting in a much slower runtime—0.6 seconds per prediction for 19 references compared to only 0.011 seconds for Eff-GRot, making our method over 50$\times$ faster.

\subsection{Runtime and Memory Analysis}
\label{subsec:runtime}

Eff-GRot is a computationally efficient model that offers a favorable trade-off between performance and inference time. In this section, we present qualitative comparisons with other models to highlight differences in computational cost. Additionally, we report GPU memory usage, which is an important consideration in practical deployment. Table~\ref{tab:runtime} provides the runtime requirements for performing rotation estimation for each method measured on a GTX~1660~Ti GPU. For methods using refinement, we provide separately the time needed for coarse rotation estimation, the time for refinement, as well as the total time when refinement is included. We assume that for every method, the references have been preprocessed, and we only measure the time needed for processing a query image and making the prediction. It should be noted that while our onboarding stage consists only of encoding the references into their corresponding embeddings, GSPose and OnePose++ require running a 3D reconstruction algorithm to obtain explicit 3D models, which introduces additional overhead.

\begin{figure}
    \centering
    \includegraphics[width=0.8\columnwidth]{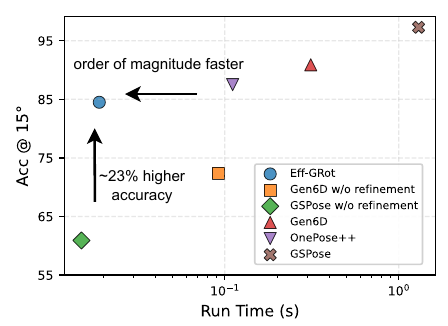}
    \caption{Performance vs runtime tradeoff. We visualize the trade-off between performance and runtime for different methods on LINEMOD. The x-axis is shown on a logarithmic scale to better capture differences in runtime. Eff-GRot exhibits a favorable balance between runtime and performance compared to baselines. }
    \label{fig:runtime}
\end{figure}

Furthermore, Figure~\ref{fig:runtime} presents an analysis of the performance-runtime tradeoff for the different methods on LINEMOD. Eff-GRot strikes a favorable balance, achieving a much more accurate rotation estimation when similar runtime is considered. The refinement steps of Gen6D \cite{liu2022gen6d} and GSPose\cite{cai2024gs} significantly impact runtime, with the differentiable rendering of GSPose being the most expensive. At the same time, our method can be competitive with OnePose++ while having an order of magnitude higher inference speed.

Table~\ref{tab:memory} reports the peak GPU memory usage (in MB) required for making a prediction with each model. Eff-GRot demonstrates significantly less memory consumption than Gen6D and GSPose. While OnePose++ achieves the lowest memory usage overall, this comes at the cost of slower inference time. For Eff-GRot, we report two values corresponding to two encoder backbones: Distilled VAE (256 MB) and SD-VAE (588 MB). The Distilled VAE offers a significant decrease in memory usage, making it a suitable alternative. Overall, Eff-GRot offers a favorable balance between memory efficiency and predictive performance.

\begin{table}[tb]
\centering
\resizebox{\columnwidth}{!}{
\begin{tabular}{lcccc}
\toprule
Model & 3D Recon. & Estimation & Refinement & Total \\
& Required & Time (s) & Time (s) & Time (s) \\
\midrule
GSPose w/o                & \xmark & 0.015 & --    & 0.015 \\
Eff-GRot                 & \xmark & 0.019 & --    & 0.019 \\
Gen6D w/o                 & \xmark & 0.092 & --    & 0.092 \\
OnePose++                 & \cmark & 0.111 & --    & 0.111 \\
Gen6D                     & \xmark & 0.092 & 0.220 & 0.312 \\
GSPose                    & \cmark & 0.015 & 1.29  & 1.30 \\
\bottomrule
\end{tabular}%
}
\caption{Runtime comparison on LINEMOD. The runtime required for a rotation prediction for different methods is presented, with "w/o" suffix indicating that the refinement step is excluded.}
\label{tab:runtime}
\end{table}

\begin{table}[tb]
\centering
\scriptsize
\renewcommand{\arraystretch}{0.85}
\setlength{\tabcolsep}{4pt}
\resizebox{\columnwidth}{!}{%
\begin{tabular}{lcccc}
\toprule
\textbf{} & Gen6D & OnePose++ & GSPose & Eff-GRot \\
\midrule
\textbf{Peak Mem. (MB)} & 705  & 201  & 897  & 256 / 588  \\
\bottomrule
\end{tabular}%
}
\caption{Peak GPU memory usage (MB). Eff-GRot: Distilled / SD-VAE.}
\label{tab:memory}
\end{table}

\subsection{Ablation Study} 
\label{subsec:ablation}

\begin{figure*}[t]
    \centering
    \includegraphics[width=0.8\textwidth]{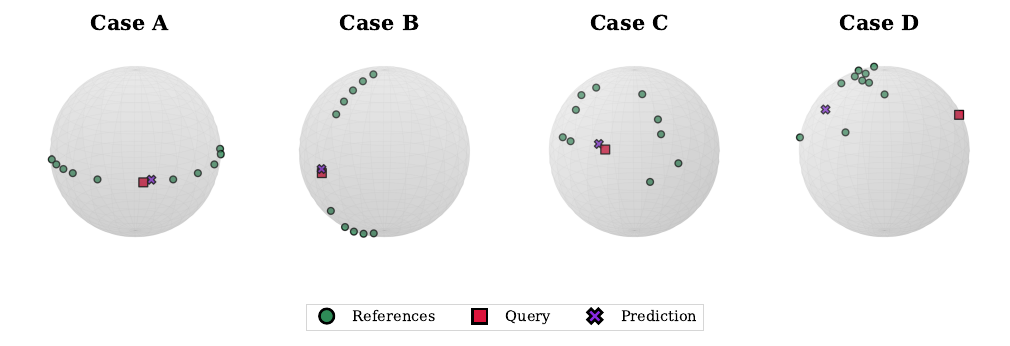}
    \caption{Visualization of rotation prediction across different reference distribution scenarios for the cat object. The predicted (purple cross) and ground truth (red square) rotations are visualized for four different reference distributions. Rotations are visualized as points on the viewing sphere, representing camera locations looking toward the object at the center of the sphere.  A,B) The references and query vary along a single axis (azimuth and elevation), showing that the model can correctly position the query between the references. C) The query falls within a region covered by the references, enabling successful interpolation for rotation prediction. D) The query lies far outside the reference coverage region, leading to a failure in rotation prediction.}
    \label{fig:interpolation}
\end{figure*}

\paragraph{\textbf{Number of References.}} 
We analyze how the number of reference views used during training affects rotation estimation performance under varying numbers of reference views at test-time. The results in Table~\ref{tab:ref_ablation} show that training with 64 reference views leads to consistently better performance compared to training with only 16 references, across all tested configurations. These findings suggest that increasing the number of training-time references improves the model’s ability to generalize across different test-time conditions.

\begin{table}[t]
\centering
\footnotesize
\begin{tabular}{lcccccc}
\toprule
Train $K$ & \multicolumn{2}{c}{LINEMOD} & \multicolumn{2}{c}{ShapeNet} & \multicolumn{2}{c}{CO3D} \\
\cmidrule(lr){2-3} \cmidrule(lr){4-5} \cmidrule(lr){6-7}
 & \multicolumn{6}{c}{\textit{Test-time references $K$}} \\
 & 32 & 64 & 32 & 64 & 9 & 19 \\
\midrule
16 & 59.2 & 62.9 & 77.8 & 81.5 & 40.2 & 63.4 \\
64 & \textbf{71.4} & \textbf{83.4} & \textbf{78.2} & \textbf{85.8} & \textbf{54.0} & \textbf{77.1} \\
\bottomrule
\end{tabular}
\caption{Ablation on number of training references \(K\). 
We report ACC @ 15\textdegree{} across datasets, using different numbers of test-time references. 
Each row corresponds to a model trained with a specific number of references \(K\). 
Image encoding is performed using the Distilled VAE. 
Training with more references (e.g., 64 vs. 16) consistently improves performance across test-time settings.}
\label{tab:ref_ablation}
\end{table}

\paragraph{\textbf{Rotation Augmentation.}} 
As described in Section~\ref{subsec:training_inference}, during training, we dynamically modify the rotation labels for each reference in a batch to prevent the network from memorizing the visual appearance of an image and directly associating it with its orientation, which would result in overfitting to the objects in the training data. This encourages the model to infer the rotation by comparing the query with the references.

In Table~\ref{tab:pose_aug}, we provide quantitative results on the effect of this transformation on the model's generalization ability. These results were obtained from a preliminary version of the model trained and tested on uniform-background images. While this differs from the main experiments (which include background augmentation during training and are tested on full-scene backgrounds), it clearly demonstrates the benefit of rotation augmentation across datasets.

As shown, the model trained with this augmentation consistently outperforms the one trained without it across different numbers of references. This validates our intuition that such transformations enable the model to perform a meaningful comparison rather than relying on visual appearance.

\begin{table}[t]
\centering
\resizebox{0.8\columnwidth}{!}{
\begin{tabular}{lccc}
\toprule
Rotation Aug. & LINEMOD & ShapeNet & CO3D \\
\midrule
\ding{55} & 69.5 & 80.8 & 56.0 \\
\ding{51} & \textbf{80.0} & \textbf{88.3} & \textbf{77.5} \\
\bottomrule
\end{tabular}
}
\caption{Effect of pose augmentation across datasets. Acc @ 15° with 64 references for LINEMOD and ShapeNet and 19 for CO3D. Pose augmentation consistently results in improved performance.}
\label{tab:pose_aug}
\end{table}

\paragraph{\textbf{Image Encoder.}} We conduct an ablation study (see Table~\ref{tab:distilled}) to compare the performance of Eff-GRot when trained with the Distilled and SD-VAE, respectively. On the LINEMOD dataset, the SD-VAE yields higher accuracy, whereas the Distilled VAE performs better on CO3D and ShapeNet, although the improvement on ShapeNet is marginal. These results indicate that the lightweight encoder achieves comparable performance while offering improved runtime efficiency, making it a practical choice for real-time applications.

\begin{table}
\centering
\renewcommand{\arraystretch}{1.2}
\resizebox{\columnwidth}{!}{
\begin{tabular}{lcccc}
\toprule
Encoder (Params) & Runtime (ms) & LINEMOD & ShapeNet & CO3D \\
\midrule
Distilled VAE (1M) & 13 & 83.6 & 94.8 & 77.1 \\ 
SD-VAE (34M) & 19 & 84.8 & 94.6 & 76.0 \\ 
\bottomrule 
\end{tabular}
}
\caption{Effect of image encoder. We report Acc@15\textdegree{} across datasets for two different image encoders, along with the runtime required for rotation prediction. The lightweight encoder serves as an efficient alternative, achieving comparable or even better performance in some cases while offering substantial efficiency gains.}

\label{tab:distilled}
\end{table}

\begin{figure*}[t]
    \centering
    \includegraphics[width=0.9\textwidth]{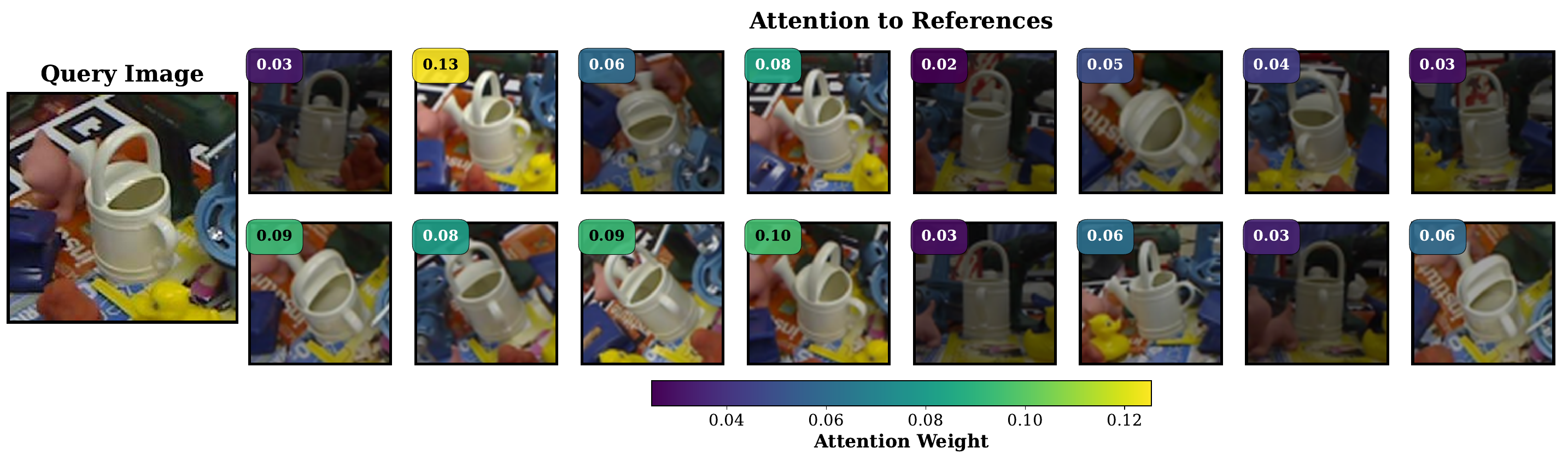}
    \caption{Transformer attention. For a given query image, we visualize the transformer's attention weights to highlight the reference images it relies on most for prediction. We use higher brightness to indicate a higher attention score, and also provide the attention values. Nearby viewpoints have higher attention values, indicating their greater influence on the final rotation estimation.}
    \label{fig:attention_to_refs}
\end{figure*}

\begin{figure}[t]
    \centering
    \includegraphics[width=\columnwidth]{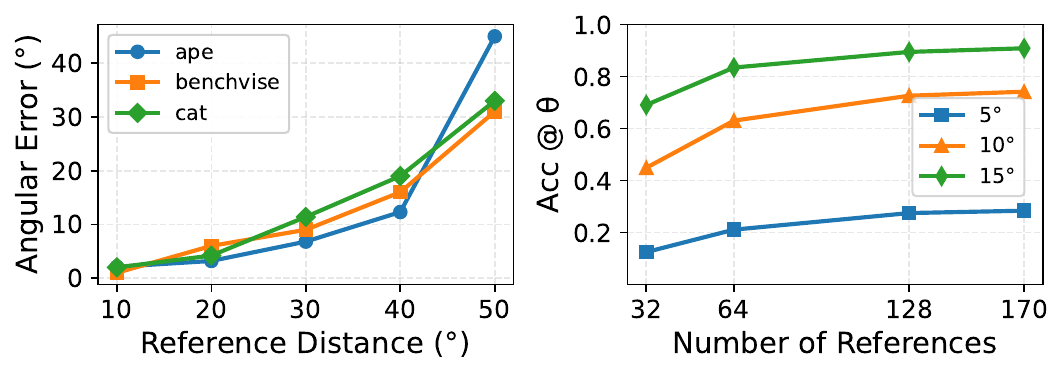}
    \caption{\textbf{Left)} Angular error versus angular separation of reference viewpoints (azimuth) for three LINEMOD objects, with the query fixed between the references. Distances shown are from the query to each reference. 
\textbf{Right)} Effect of reference count on performance across accuracy thresholds on LINEMOD.}
    \label{fig:error_plot}
\end{figure}

\subsection{Further Analysis}
\label{subsec:qualitative}
We present visualizations to provide deeper insight into our model’s behavior. 
Figure~\ref{fig:interpolation} analyzes interpolation capabilities and failure cases using synthetic data from LINEMOD objects. 
In the first and second images, where the rotation varies along a single axis (elevation and azimuth, respectively), the model accurately interpolates and predicts the correct rotation. 
The third image shows a query within the coverage region of upper-hemisphere references, again yielding accurate predictions. 
The last image highlights a failure case where the query lies outside the coverage region, making accurate prediction difficult. 
These visualizations indicate that the model learns a form of rotation-aware interpolation in the latent space.

We further analyze the model’s behavior through attention visualizations. 
In Figure~\ref{fig:attention_to_refs}, we show attention scores from the last transformer layer over reference images for a LINEMOD object. 
The model consistently attends to references with rotations closest to the query, suggesting it relies on local interpolation to infer rotation.

\begin{figure}[htb]
    \centering
    \includegraphics[width=\columnwidth]{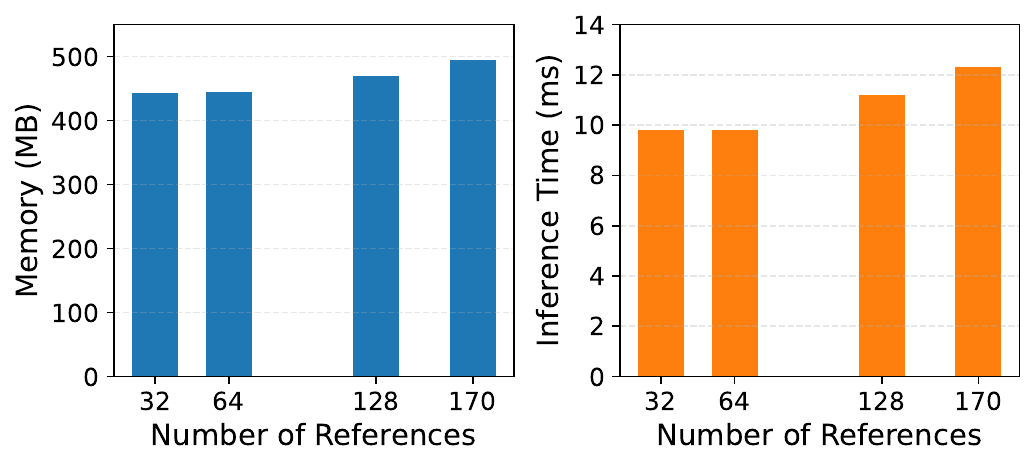}
    \caption{
\textbf{Left)} GPU memory usage versus reference count. 
\textbf{Right)} Inference time versus reference count. 
GPU memory and inference time increase only marginally with more references.
}
    \label{fig:mem_inf_bars}
\end{figure}

We also examine how the number of reference images affects performance across different thresholds and its impact on GPU memory usage and inference time. 
Increasing the number of references consistently improves performance. 
As shown in Figure~\ref{fig:error_plot}, when the threshold is tightened from $15^\circ$ to $10^\circ$, the model still maintains reasonable performance, but its accuracy drops more markedly at $5^\circ$.
Figure~\ref{fig:mem_inf_bars} indicates that both inference time and memory usage exhibit a small increase with more references, suggesting that additional views can enhance performance without substantially affecting efficiency.

Finally, we assess how the angular separation between a query and its reference views influences rotation accuracy. Using synthetic data from three LINEMOD objects, we vary only the azimuth angle and place the query between two reference views whose angular offsets increase from  $10^\circ$ to $50^\circ$. 
The left plot of Figure~\ref{fig:error_plot} shows that the error remains relatively low up to $30^\circ$--$40^\circ$ (depending on the object) but increases sharply at $50^\circ$, confirming that the model interpolates effectively between nearby viewpoints but struggles with larger gaps.

\section{Conclusion}

This work introduced Eff-GRot, a generalizable rotation estimation approach that directly predicts the rotation of novel objects in a single forward pass using only RGB inputs. Leveraging a transformer-based architecture, our model performs rotation-aware comparisons in the latent space, eliminating the need for multi-stage refinement. Experiments on both synthetic and real-world datasets show that Eff-GRot strikes an effective balance between runtime efficiency and predictive accuracy, achieving competitive rotation estimation performance even compared to slower methods. The architecture is simple, scalable, and efficient, making it well-suited for latency-sensitive applications.

\paragraph{Limitations.} Eff-GRot can struggle with large rotation variations, particularly when the reference set fails to cover the range of possible viewpoints sufficiently. In such cases, the model's prediction accuracy may degrade due to limited interpolation capability in the latent space. Furthermore, our approach is not well-suited for applications requiring fine-grained rotation alignment, as it prioritizes efficiency over accuracy. In future work, we aim to enhance robustness under sparse reference coverage and explore lightweight refinement strategies that preserve real-time performance.

\clearpage
\section*{Acknowledgment}
This research was supported by the C-DATA project: ‘Digitalizing AM through data sharing and AI’ (HBC.2022.0135), funded by VLAIO (Flanders Innovation \& Entrepreneurship).

{
    \small
    \bibliographystyle{ieeenat_fullname}
    \bibliography{main}
}

\end{document}